\documentclass[
 two-column
]{ceurart}

\usepackage{listings}
\usepackage{amsmath}
\begin{document}

\copyrightyear{2022}
\copyrightclause{Copyright for this paper by its authors.
  Use permitted under Creative Commons License Attribution 4.0
  International (CC BY 4.0).}

\conference{Forum for Information Retrieval Evaluation, December 12-15, 2024, India}

\title{YouTube Comments Decoded: Leveraging LLMs for Low Resource Language Classification}


\author[1]{Aniket Deroy}[%
orcid=0000-0001-7190-5040,
email=roydanik18@kgpian.iitkgp.ac.in,
%url=https://yamadharma.github.io/,
]
\cormark[1]
\fnmark[1]
\address[1]{IIT Kharagpur,
  Kharagpur, India}
%\address[2]{}%Joint Institute for Nuclear Research,
  %6 Joliot-Curie, Dubna, Moscow region, 141980, Russian Federation}

\author[1]{Subhankar Maity}[%
orcid=0009-0001-1358-9534,
email=subhankar.ai@kgpian.iitkgp.ac.in,
%url=https://kmitd.github.io/ilaria/,
]
%\address[3]{Vrije Universiteit Amsterdam, De Boelelaan 1105, 1081 HV Amsterdam, The Netherlands}

\cortext[1]{Corresponding author.}
%\fntext[1]{These authors contributed equally.}

\begin{abstract}
Sarcasm detection is a significant challenge in sentiment analysis, particularly due to its nature of conveying opinions where the intended meaning deviates from the literal expression. This challenge is heightened in social media contexts where code-mixing, especially in Dravidian languages, is prevalent. Code-mixing involves the blending of multiple languages within a single utterance, often with non-native scripts, complicating the task for systems trained on monolingual data. This shared task introduces a novel gold standard corpus designed for sarcasm and sentiment detection within code-mixed texts, specifically in Tamil-English and Malayalam-English languages.
%Tamil, spoken widely in India, Sri Lanka, and among the global Tamil diaspora, has a rich scriptural heritage with 12 vowels, 18 consonants, and additional characters used by minority languages. Malayalam, predominantly spoken in Kerala, India, utilizes an alpha-syllabic script that is a hybrid of alphabetic and syllable-based systems. However, social media users frequently employ the Roman script due to its ease of input, resulting in a predominance of code-mixed data for these under-resourced languages on digital platforms.
The primary objective of this task is to identify sarcasm and sentiment polarity within a code-mixed dataset of Tamil-English and Malayalam-English comments and posts collected from social media platforms. Each comment or post is annotated at the message level for sentiment polarity, with particular attention to the challenges posed by class imbalance, reflecting real-world scenarios.% Participants in this task will receive development, training, and test datasets to classify whether a given YouTube comment is sarcastic or not. This initiative aims to advance research into the nuances of sarcasm expression in code-mixed social media contexts, contributing to more robust sentiment analysis systems for multilingual environments.
In this work, we experiment with state-of-the-art large language models like GPT-3.5 Turbo via prompting to classify comments into sarcastic or non-sarcastic categories. We obtained a macro-F1 score of 0.61 for Tamil language. We obtained a macro-F1 score of 0.50 for Malayalam language.

\end{abstract}


\maketitle

\section{Introduction}
Sarcasm, a form of expression where the intended meaning sharply diverges from the literal meaning, poses a significant challenge for sentiment analysis systems \cite{dr1}. This complexity arises because sarcasm often involves subtle linguistic cues and context-dependent interpretation, making it difficult for automated systems to accurately detect and classify \cite{dr2, dr3}. %The challenge is further compounded in multilingual communities where code-mixing—the practice of alternating between two or more languages within a single conversation—is prevalent, especially on social media platforms \cite{dr4}.

Sarcasm is a sharp, often humorous form of verbal irony where someone says the opposite of what they truly mean, typically to convey disdain or mockery. It relies heavily on tone and context, as the delivery can transform a seemingly straightforward statement into a biting comment\cite{dr2, dr3}. While it can be a playful way to highlight absurdities or critique behavior, sarcasm can also lead to misunderstandings, especially in written communication where vocal inflections are absent. Ultimately, it’s a double-edged sword that can entertain or offend, depending on the audience and the intent behind it.

Overall, the study of code-mixing in NLP is an evolving field with ongoing research aimed at improving the performance of various NLP tasks in multilingual contexts. As multilingualism continues to grow globally, the development of models that can effectively handle code-mixed data is becoming increasingly important.

Code-mixing~\cite{dr5} in NLP refers to the blending of two or more languages within a single sentence or conversation, commonly seen in multilingual societies. It poses unique challenges for tasks like language identification, part-of-speech tagging, and sentiment analysis. Traditional NLP models often struggle with code-mixed data due to the complex interaction between languages. To address this, researchers have developed advanced neural network models, such as RNNs and CNNs, that better capture linguistic context. Code-mixed text also complicates machine translation and speech recognition, requiring specialized approaches to handle language switching effectively. As multilingualism becomes more prevalent, the importance of developing robust NLP systems capable of processing code-mixed data continues to grow.

Code-mixing is a widespread phenomenon in multilingual societies, particularly in Dravidian languages such as Tamil and Malayalam, where users frequently switch between native languages and English \cite{dr5}. These code-mixed texts often appear in non-native scripts, such as Roman, due to the convenience of typing on digital platforms \cite{dr5}. This creates a unique set of challenges for natural language processing (NLP) systems, which are typically trained on monolingual data and thus struggle to handle the intricacies of code-switching across different linguistic levels, such as syntax, semantics, and phonetics \cite{dr6}.

The Dravidian languages, with their rich cultural and linguistic history, present a distinct case for code-mixed text analysis \cite{dr7}. Tamil, with its official status in India, Sri Lanka, and Singapore, and a significant diaspora community, has a script that evolved from ancient scripts like Tamili and the Chola-Pallava script \cite{dr8}. Similarly, Malayalam, predominantly spoken in Kerala, India, employs an alpha-syllabic script that blends alphabetic and syllable-based writing systems \cite{dr9}. However, on social media, these languages are often typed using the Roman script, contributing to a growing corpus of code-mixed data.

%Given the increasing use of code-mixed languages on social media, there is a pressing need for robust systems capable of detecting sarcasm and analyzing sentiment in these contexts \cite{dr10}. Traditional NLP systems, trained on monolingual and monolingual-script data, often fail to accurately interpret code-mixed texts due to the linguistic complexity and variability inherent in code-switching \cite{dr11}.

To address these challenges, this paper introduces a prompt-based method \cite{cm25} using GPT-3.5 Turbo \cite{cm24} for sarcasm and sentiment detection in code-mixed texts, specifically focusing on Tamil-English and Malayalam-English language pairs. This shared task~\cite{chakravarthi2021findings,chakravarthi2022can,chakravarthi2022hope,chakravarthi2023overview,chakravarthi2023sarcasm,sripriya2023findings} aims to advance research in sentiment analysis by providing a dataset that reflects real-world scenarios where code-mixing is prevalent and by encouraging the development of systems that can accurately classify sarcasm and sentiment polarity in these under-resourced languages.

Participants in this task were provided development, training, and test datasets composed of comments and posts collected from social media platforms such as YouTube.% Each comment or post is annotated for sentiment polarity at the message level, with an emphasis on detecting sarcasm. The task not only highlights the linguistic challenges posed by code-mixing but also addresses the issue of class imbalance, which mirrors the complexities of sentiment distribution in real-world social media data.

%This work is a significant step toward improving the accuracy of sentiment analysis systems in multilingual and code-mixed environments, with broader implications for the development of NLP tools that can handle the diverse and evolving linguistic landscape of social media.

In this work, we explore the capabilities of state-of-the-art large language models, specifically GPT-3.5 Turbo \cite{cm24}, by leveraging prompt-based techniques to classify social media comments into sarcastic or non-sarcastic categories. GPT-3.5 Turbo \cite{cm24}, a variant of OpenAI's GPT-3.5 model, is known for its advanced natural language understanding and generation capabilities, making it a promising candidate for handling the nuanced task of sarcasm detection, particularly in code-mixed texts.

Our team, TextTitans, participated in a competition aimed at advancing sarcasm detection in Tamil and Malayalam. The results of our systems' performance provide valuable insights into the current state of NLP for these languages. For Tamil, our system achieved a macro-F1 score of 0.61, placing us 9th among competing teams. The macro-F1 score, which balances precision and recall across multiple classes, reflects our system's ability to effectively handle the nuances of Tamil sarcasm. However, despite the relatively strong performance indicated by the F1 score, we achieved 9th rank. %suggests that the competition was particularly fierce, with other systems either matching or surpassing our performance.

In contrast, our system's performance for Malayalam resulted in a macro-F1 score of 0.50, securing the 13th position in the rankings. This score, while moderate, highlights the challenges inherent in processing Malayalam text, particularly in capturing the subtleties of sarcastic expressions. The lower rank compared to Tamil indicates that our system faced greater difficulties with Malayalam, and that several other systems were more adept at handling these challenges.

These results underscore the varying levels of difficulty and competition present in sarcasm detection for Tamil and Malayalam. The higher performance in Tamil suggests that our system is more attuned to the linguistic characteristics of Tamil sarcasm, while the more modest performance in Malayalam points to areas where further research and refinement are necessary. This study not only contributes to the growing body of work in multilingual sarcasm detection but also highlights the importance of continuing to develop robust NLP systems capable of addressing the complexities of underrepresented languages.
\section{Related Work}

Sarcasm detection has been a topic of significant interest in the field of natural language processing (NLP) due to its complex and context-dependent nature \cite{dr12}. Traditional approaches to sarcasm detection primarily relied on feature-based methods, where linguistic cues such as punctuation, lexical patterns, and syntactic structures were used to identify sarcasm \cite{dr12}. For instance, works by \cite{dr13} and \cite{dr14} utilized syntactic patterns and hashtag-based supervision to improve sarcasm detection in English texts. These approaches, however, often struggled with generalization, particularly in scenarios involving nuanced and culturally specific forms of sarcasm.

%With the advent of deep learning, more sophisticated models, such as convolutional neural networks (CNNs) and recurrent neural networks (RNNs), were employed to capture the intricate patterns in text that signify sarcasm \cite{dr15}.
\cite{dr16} introduced an approach using deep neural networks that leveraged context to improve sarcasm detection. Similarly, \cite{dr17} explored attention-based models that better captured the dependencies within a sentence that could indicate sarcasm. These deep learning models significantly advanced the state-of-the-art in sarcasm detection, particularly in monolingual contexts.

%Code-mixing in Natural Language Processing (NLP) involves the blending of two or more languages within a single conversation, sentence, or even word. This phenomenon is prevalent in multilingual societies and presents unique challenges for various NLP tasks, including machine translation, sentiment analysis, language modeling, and speech recognition.

One of the primary challenges in dealing with code-mixed text is language identification~\cite{hidayatullah2022systematic}, which involves determining the language of each word or phrase within a mixed-language sentence. Traditional methods for this task often relied on n-gram models, dictionary-based approaches, and supervised learning techniques. However, these methods sometimes struggle with the complex linguistic patterns present in code-mixed data. %To address these limitations, more advanced approaches have been developed, leveraging neural networks such as Recurrent Neural Networks (RNNs) and Convolutional Neural Networks (CNNs).
These models are particularly effective at capturing the contextual information necessary to accurately identify languages in code-mixed text.

Another critical area of research is part-of-speech tagging~\cite{sequiera2015pos} for code-mixed sentences. This task becomes more complicated due to the interaction between different languages' grammatical structures. Researchers have employed techniques like Conditional Random Fields (CRFs) and sequence-to-sequence models, adapting them to handle the intricacies of mixed-language data. The use of multilingual embeddings, where words from different languages are mapped into a shared vector space, has also been shown to improve the accuracy of part-of-speech tagging in code-mixed scenarios.

In the context of sentiment analysis~\cite{chakravarthi2020overview}, code-mixing introduces additional complexity due to the potential for differing sentiment expressions across languages. Traditional sentiment analysis tools often fail when applied directly to code-mixed text. To overcome this, researchers have explored methods such as using bilingual lexicons, translating the text into a single language before analysis, and employing deep learning models that can process code-mixed input directly.

Machine translation~\cite{jawahar2021exploring} for code-mixed text is another challenging area. Standard machine translation systems are usually trained on monolingual data, making them less effective at handling code-mixed sentences. Researchers have proposed various strategies to address this, including the use of synthetic code-mixed data for training, as well as incorporating language tags and contextual embeddings into translation models. These approaches aim to create more robust translation systems capable of handling the nuances of code-mixing.

Additionally, speech recognition~\cite{long2020acoustic} in code-mixed environments requires models that can accurately transcribe spoken language that switches between different languages. This task is particularly challenging because traditional speech recognition systems are typically designed for single-language input. Recent advances involve the use of end-to-end models, such as attention-based encoder-decoder architectures, which have shown promise in recognizing and transcribing code-mixed speech.

Our contribution is situated at the intersection of sarcasm detection, code-mixed text processing, and the application of large language models, addressing a critical gap in the literature and advancing the state-of-the-art in this complex domain.

\section{Dataset}

%The Tamil training dataset is 29570 comments. The Kannada Training dataset is 13188 comments.
The Tamil testing dataset is 6338 comments. The Malayalam testing dataset is 2826 comments.
Since we used effectively only the test dataset for our predictions, we only mention the statistics corresponding to the test dataset.

Given that our evaluation focused entirely on the test datasets, the statistics mentioned correspond to these specific datasets. The decision to use only the test datasets for prediction ensures that the performance metrics, such as the macro-F1 scores, accurately reflect the model's generalization capabilities on unseen data.

The Tamil testing dataset, with its larger size of 6,338 comments, provides a broad spectrum of inputs for the model to process, potentially leading to more robust insights into the model's strengths and weaknesses. On the other hand, the smaller Malayalam testing dataset, with 2,826 comments, presents its own set of challenges, particularly in a low-resource language context where data is often scarcer.

By focusing on these testing datasets, our analysis remains rooted in real-world applicability, offering a clear view of how well the models perform in practical scenarios without additional interventions or training phases.

\section{Task Definition}
%Objective:
The task is to accurately classify YouTube comments that are written in Tamil-English and Malayalam-English (code-mixed) into two distinct categories: Sarcastic and Non-sarcastic, using a AI model.

\section{Methodology}

Prompting \cite{cm25} is an effective approach for solving the problem of sarcasm detection in code-mixed texts, particularly in under-resourced languages like Tamil-English and Malayalam-English, for several reasons:

\begin{itemize}[-]

\item \textbf{Natural Language Understanding:}
Prompting allows the model to interpret and execute tasks based on human-like instructions. By framing the classification task as a natural language prompt~\cite{allen1988natural}, the model can utilize its language understanding capabilities to classify text according to the specified categories.
The prompt provides context to the model, guiding it to focus on specific aspects of the input (e.g., determining sarcasm). This ensures that the model’s response is aligned with the task at hand.
    
\item \textbf{Leverage of Pretrained Knowledge:} Large language models like GPT-3.5 Turbo have been trained on vast amounts of text data, encompassing diverse linguistic patterns, cultural contexts, and language nuances \cite{dr25}. Prompting allows us to tap into this rich, pre-trained knowledge base, enabling the model to interpret sarcasm even in challenging code-mixed scenarios \cite{cm25}. By carefully crafting prompts, we can guide the model to focus on specific aspects of the input, such as tone, context, and language switching, which are critical for sarcasm detection.

\item \textbf{Reduced Need for Fine-Tuning:} With the right prompt, large pre-trained models (like GPT)~\cite{colon2021combining} can perform various classification tasks without the need for extensive fine-tuning on task-specific data. This is particularly useful when data or computational resources for fine-tuning are limited. Prompts leverage the model’s existing knowledge, gained from vast pre-training on diverse text corpora, enabling it to generalize to new tasks with minimal additional training.

\item \textbf{Adaptability to Multilingual and Code-Mixed Texts:} Traditional machine learning models often require large, labeled datasets for each specific task and language, which are scarce for many code-mixed languages \cite{dr26}. In contrast, prompting large language models provides a flexible way to adapt to different languages and dialects, even when training data is limited \cite{cm25}. The model's inherent ability to understand and generate text in multiple languages, including code-mixed varieties, makes prompting an attractive approach for this task.

\item \textbf{Improved Interpretability:}
The explicit nature of prompts makes it easier to understand how the model is interpreting the task~\cite{jie2024interpretable}, as the instructions are directly embedded in the input. This transparency can help in analyzing and improving the model’s performance. By examining the model’s responses to different prompts, researchers can gain insights into where and why the model might be making classification errors, leading to more targeted improvements.

\item \textbf{Reduced Need for Task-Specific Data:} Building a model from scratch or fine-tuning a model for sarcasm detection in code-mixed texts would typically require a substantial amount of labeled data, which is often difficult to obtain for under-resourced languages \cite{dr27}. Prompting, however, minimizes the need for such extensive labeled datasets, as it allows the model to utilize its general language understanding capabilities with minimal task-specific adjustments \cite{cm25}. This makes it a more feasible and efficient solution for resource-constrained settings.

\item \textbf{Flexibility and Rapid Experimentation:} Prompting enables rapid experimentation with different strategies to optimize model performance \cite{cm25}. Researchers can quickly test various prompt formulations to see which one elicits the most accurate and context-aware responses from the model. This flexibility is particularly beneficial for complex tasks like sarcasm detection, where the model's interpretation can vary significantly depending on how the problem is framed.

\item \textbf{Scalability Across Different Scenarios:} Once an effective prompting strategy is developed, it can be easily adapted or scaled to other similar tasks or languages without the need for extensive retraining \cite{dr28, cm25}. This is particularly useful in multilingual environments where a single model might need to handle multiple languages or dialects. Prompting allows for the reuse of the same model across different linguistic contexts with minimal modifications.

\end{itemize}
In summary, prompting is a powerful and efficient approach to tackle the problem of sarcasm detection in code-mixed texts, offering the benefits of leveraging large-scale pretrained models, adapting to multilingual and code-mixed scenarios, reducing dependency on large annotated datasets, enabling rapid experimentation, and providing scalability across different linguistic contexts.

\subsection{Prompt Engineering-Based Approach}
We used the GPT-3.5 Turbo model via prompting to solve the classification task.
We used GPT-3.5 Turbo in zero-Shot mode via prompting.
After the prompt is provided to the LLM, the following steps happen internal to the LLM while generating the output. The following outlines the steps~\cite{cm24,vaswani2017attention,radford2018improving} that occur internally within the LLM, summarizing the prompting approach using GPT-3.5 Turbo:\\

\textbf{Step 1: Tokenization}

\begin{itemize}
    \item \textbf{Prompt:} \( X = [x_1, x_2, \dots, x_n] \)
    \item The input text (prompt) is first tokenized into smaller units called tokens. These tokens are often subwords or characters, depending on the model's design.
    \item \textbf{Tokenized Input:} \( T = [t_1, t_2, \dots, t_m] \)
\end{itemize}

\textbf{Step 2: Embedding}

\begin{itemize}
    \item Each token is converted into a high-dimensional vector (embedding) using an embedding matrix \( E \).
    \item \textbf{Embedding Matrix:} \( E \in \mathbb{R}^{|V| \times d} \), where \( |V| \) is the size of the vocabulary and \( d \) is the embedding dimension.
    \item \textbf{Embedded Tokens:} \( T_{\text{emb}} = [E(t_1), E(t_2), \dots, E(t_m)] \)
\end{itemize}

\textbf{Step 3: Positional Encoding}

\begin{itemize}
    \item Since the model processes sequences, it adds positional information to the embeddings to capture the order of tokens.
    \item \textbf{Positional Encoding:} \( P(t_i) \)
    \item \textbf{Input to the Model:} \( Z = T_{\text{emb}} + P \)
\end{itemize}

\textbf{Step 4: Attention Mechanism (Transformer Architecture)}

\begin{itemize}
    \item \textbf{Attention Score Calculation:} The model computes attention scores to determine the importance of each token relative to others in the sequence.
    \item \textbf{Attention Formula:}
    \begin{equation}
    \text{Attention}(Q, K, V) = \text{softmax}\left(\frac{QK^T}{\sqrt{d_k}}\right)V
    \end{equation}
    \item where \( Q \) (query), \( K \) (key), and \( V \) (value) are linear transformations of the input \( Z \).
    \item This attention mechanism is applied multiple times through multi-head attention, allowing the model to focus on different parts of the sequence simultaneously.
\end{itemize}

\textbf{Step 5: Feedforward Neural Networks}

\begin{itemize}
    \item The output of the attention mechanism is passed through feedforward neural networks, which apply non-linear transformations.
    \item \textbf{Feedforward Layer:}
    \begin{equation}
    \text{FFN}(x) = \max(0, xW_1 + b_1)W_2 + b_2
    \end{equation}
    \item where \( W_1, W_2 \) are weight matrices and \( b_1, b_2 \) are biases.
\end{itemize}

\textbf{Step 6: Stacking Layers}

\begin{itemize}
    \item Multiple layers of attention and feedforward networks are stacked, each with its own set of parameters. This forms the "deep" in deep learning.
    \item \textbf{Layer Output:}
    \begin{equation}
    H^{(l)} = \text{LayerNorm}(Z^{(l)} + \text{Attention}(Q^{(l)}, K^{(l)}, V^{(l)}))
    \end{equation}
    \begin{equation}
    Z^{(l+1)} = \text{LayerNorm}(H^{(l)} + \text{FFN}(H^{(l)}))
    \end{equation}
\end{itemize}

\textbf{Step 7: Output Generation}

\begin{itemize}
    \item The final output of the stacked layers is a sequence of vectors.
    \item These vectors are projected back into the token space using a softmax layer to predict the next token or word in the sequence.
    \item \textbf{Softmax Function:}
    \begin{equation}
    P(y_i|X) = \frac{\exp(Z_i)}{\sum_{j=1}^{|V|} \exp(Z_j)}
    \end{equation}
    \item where \( Z_i \) is the logit corresponding to token \( i \) in the vocabulary.
    \item The model generates the next token in the sequence based on the probability distribution, and the process repeats until the end of the output sequence is reached.
\end{itemize}

\textbf{Step 8: Decoding}

\begin{itemize}
    \item The predicted tokens are then decoded back into text, forming the final output.
    \item \textbf{Output Text:} \( Y = [y_1, y_2, \dots, y_k] \)
\end{itemize}

\begin{figure}[h!]
  \centering
  \includegraphics[width=0.55\linewidth]{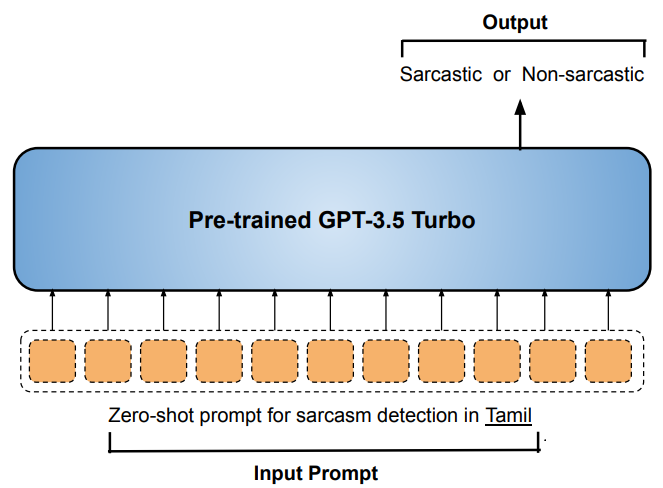}
  \caption{Architechture diagram for sarcasm detection in Tamil.} \label{fig1}
  %\Description{A woman and a girl in white dresses sit in an open car.}
\end{figure}

We used the following prompt for Tamil language for the purpose of classification: "\textit{Please Check whether the comment-<Text> is Sarcastic or Non-sarcastic. Only state Sarcastic or Non-sarcastic}".
The figure representing the methodology is shown in Figure ~\ref{fig1}.

\begin{figure}[h!]
  \centering
  \includegraphics[width=0.55\linewidth]{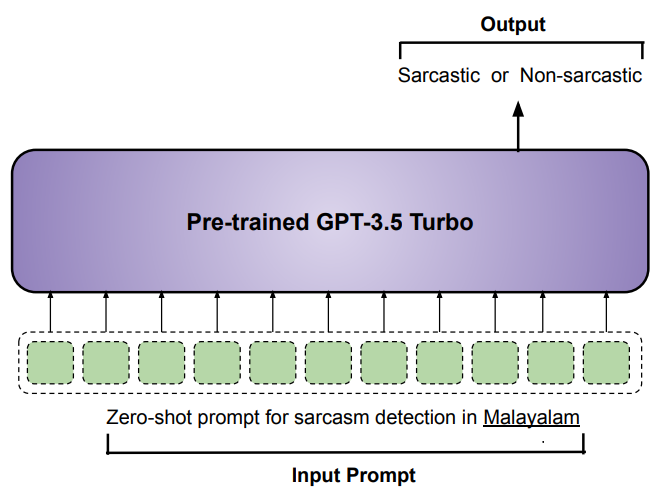}
  \caption{Architechture diagram for sarcasm detection in Malayalam.} \label{fig2}
  %\Description{A woman and a girl in white dresses sit in an open car.}
\end{figure}

We used the following prompt for Malayalam language for the purpose of classification: "\textit{Please Check whether the comment-<Text> is Sarcastic or Non-sarcastic. Only state Sarcastic or Non-sarcastic}".
The figure representing the methodology is shown in Figure ~\ref{fig2}.
We run the GPT model at temperatures of 0.7, 0.8, and 0.9.

We have used two different figures (Figure~\ref{fig1} and Figure~\ref{fig2}) for the two sarcasm detection models in Tamil and Malayalam.

The methodology employed for the classification task involved using language-specific prompts to determine whether a comment is sarcastic or non-sarcastic, with this approach applied to both Tamil and Malayalam languages. For Tamil, the prompt used was: “Please Check whether the comment-<Text> is Sarcastic or Non-sarcastic. Only state Sarcastic or Non-sarcastic”. This prompt was carefully crafted to direct the model's focus solely on the content of the comment, ensuring that the output would be either "Sarcastic" or "Non-sarcastic" without any additional context or information.

Similarly, for Malayalam, the prompt used was: “Please Check whether the comment-<Text> is Sarcastic or Non-sarcastic. Only state Sarcastic or Non-sarcastic”. This prompt, is similar in structure to the Tamil prompt, was designed to maintain consistency in the classification task across both languages.

The process began with the preparation of input, where comments in Tamil and Malayalam were fed into the model using their respective prompts. The model then processed each comment according to the instructions provided in the prompt, returning a binary classification of either "Sarcastic" or "Non-sarcastic" for each comment.

%Subsequently, the outputs from the model were analyzed to assess the accuracy of sarcasm detection in both languages. Any patterns of misclassification were identified for potential refinement of the model or the prompts. Additionally, the classification results for Tamil and Malayalam were compared to identify any language-specific challenges or advantages in sarcasm detection. The methodologies for Tamil and Malayalam were visually represented in figures (Figure ~\ref{fig1} and Figure ~\ref{fig2}, respectively), highlighting any differences or similarities in the approach used for each language.

In conclusion, this methodology ensured a consistent classification task across different languages while accommodating the unique characteristics of Tamil and Malayalam. The use of clear and concise prompts enabled the model to focus on the core task of sarcasm detection, leading to reliable and interpretable results.

\section{Results}
Our team named \textbf{TextTitans}~\cite{sarcasm2024overview}, for the Tamil language the macro-F1 score is 0.61 with a rank of 9th. For Malayalam language the macro-F1 score is 0.50 with a rank of 13th.

For Malayalam, the system has achieved a macro-F1 score of 0.50, which represents the harmonic mean of precision and recall across multiple classes, treating each class equally.
This score places the system in the 13th position among other competing models or systems. A rank of 13th suggests that while the performance is moderate, there are several other systems that have performed better.

For Tamil, the system has a macro-F1 score of 0.61, indicating better performance than the Malayalam system in terms of the balance between precision and recall.
However, despite the higher F1 score, this system is ranked 9th. This lower rank, despite a better score, suggests that the competition for Tamil language tasks was fiercer, with more systems achieving higher or comparable performance levels.

\begin{table}[h!]
\centering
\renewcommand{\arraystretch}{1}
\scalebox{0.65}{
\begin{tabular}{|l|c|c|c|c|}
\hline
\textbf{} & \textbf{Precision} & \textbf{Recall} & \textbf{F1-Score} & \textbf{Support} \\ \hline
\textbf{Non-sarcastic} & 0.82 & 0.73 & 0.77 & 2314 \\ \hline
\textbf{Sarcastic} & 0.18 & 0.27 & 0.22 & 512 \\ \hline
\textbf{Micro avg} & 0.65 & 0.65 & 0.65 & 2826 \\ \hline
\textbf{Macro avg} & 0.50 & 0.50 & 0.50 & 2826 \\ \hline
\textbf{Weighted avg} & 0.70 & 0.65 & 0.67 & 2826 \\ \hline
\end{tabular}
}
\caption{Classification Report for Sarcasm Detection for Malayalam language.}
\label{tab:classification_report1}
\end{table}

%\scalebox{0.5}{
\begin{table}[h!]

\centering
\renewcommand{\arraystretch}{1}
\scalebox{0.65}{
\begin{tabular}{|l|c|c|c|c|}
\hline

\textbf{} & \textbf{Precision} & \textbf{Recall} & \textbf{F1-Score} & \textbf{Support} \\ \hline
\textbf{Non-sarcastic} & 0.79 & 0.79 & 0.79 & 4621 \\ \hline
\textbf{Sarcastic} & 0.43 & 0.43 & 0.43 & 1717 \\ \hline
\textbf{Micro avg} & 0.69 & 0.69 & 0.69 & 6338 \\ \hline
\textbf{Macro avg} & 0.61 & 0.61 & 0.61 & 6338 \\ \hline
\textbf{Weighted avg} & 0.69 & 0.69 & 0.69 & 6338 \\ \hline

\end{tabular}
}
\caption{Classification Report for Sarcasm Detection for Tamil language.}
\label{tab:classification_report2}

\end{table}
%}

Table~\ref{tab:classification_report1} shows the classification report for Sarcasm Detection for Malayalam language. The model performs well with a precision of 0.82, indicating that most of the time when it predicts a comment as non-sarcastic, it is correct. However, with a recall of 0.73, it misses some non-sarcastic comments, leading to a moderately high F1-score of 0.77. The model struggles with sarcasm detection, evident by a low precision (0.18) and recall (0.27). This indicates the model has difficulty both identifying sarcastic comments and correctly predicting them, resulting in a low F1-score of 0.22. The overall performance across both classes is quite balanced (0.50 for precision, recall, and F1-score), showing the disparity in performance between the two classes. Weighted by the number of instances in each class, this metric gives a better sense of the model's performance on the dataset as a whole, showing the skew towards the Non-sarcastic class due to its larger support.

Table~\ref{tab:classification_report2} shows the classification report for Sarcasm Detection for Tamil language. The model shows strong performance with a precision, recall, and F1-score all at 0.79, indicating balanced detection and prediction of non-sarcastic comments. While still less effective at detecting sarcasm, the model has improved significantly from the first table, with a precision, recall, and F1-score of 0.43. This shows better balance in detecting sarcasm compared to the first report. The macro average has improved to 0.61 for all three metrics, indicating better overall performance across both classes, even though the model is still biased towards the Non-sarcastic class. Consistent across precision, recall, and F1-score at 0.69, showing that the model's performance is more balanced in this report compared to the first one, likely due to better handling of the Sarcastic class.

\textbf{Error Analysis:} According to Table \ref{tab:classification_report1}, the sarcasm detection results for the Malayalam language using GPT-3.5 reveal significant performance gaps, particularly in identifying sarcastic remarks. The model's low precision of 0.18 for the sarcastic class indicates a high rate of false positives, suggesting that non-sarcastic comments are often misclassified as sarcastic. Additionally, a recall of 0.27 shows that many actual sarcastic instances go undetected, resulting in a low F1-score of 0.22. In contrast, the non-sarcastic class performs better, with a precision of 0.82 and an F1-score of 0.77, reflecting the model's stronger ability to identify straightforward statements. This imbalance highlights the model's struggle with the nuances of sarcasm, indicating the need for further refinement in training strategies or additional data to enhance performance in Malayalam.

Similarly, the sarcasm detection results for the Tamil language (See Table \ref{tab:classification_report2}) indicate critical areas for improvement. The model's precision for sarcastic comments is only 0.43, which signifies a considerable number of false positives, where non-sarcastic statements are misclassified as sarcastic. The recall for the sarcastic class is also low at 0.43, suggesting that the model fails to identify many genuine sarcastic instances, leading to an F1-score of 0.43. Conversely, the non-sarcastic class shows stronger performance, achieving a precision, recall, and F1-score of 0.79. This indicates effective detection of straightforward statements. The overall balanced performance, with a macro average F1-score of 0.61, highlights the inconsistency in sarcasm detection. These findings underscore the necessity for improved training methods or more diverse datasets to bolster the model's capability to recognize sarcasm in Tamil more effectively.

\section{Conclusion}
%Conclusion
%In this paper, we addressed the challenging task of sarcasm detection in code-mixed texts, focusing on Tamil-English and Malayalam-English language pairs commonly found on social media platforms.
The complexity of code-switching, coupled with the nuanced nature of sarcasm, presents significant hurdles for traditional sentiment analysis systems, particularly in under-resourced languages.
To tackle these challenges, we experimented with state-of-the-art large language models like GPT-3.5 Turbo, leveraging prompt-based techniques to classify comments and posts as sarcastic or non-sarcastic. Our approach demonstrated the potential of prompting as an effective method for sarcasm detection in multilingual and code-mixed environments, allowing us to utilize the extensive pretrained knowledge of the model with minimal task-specific data.
The presence of a new gold standard corpus for Tamil-English and Malayalam-English code-mixed text contributes to the advancement of research in this area, providing a valuable resource for the development and evaluation of sarcasm detection systems. %Our findings highlight the importance of understanding the linguistic and contextual subtleties in code-mixed languages and underscore the need for further research to enhance the robustness and accuracy of NLP systems in such complex settings.
Overall, our work underscores the viability of prompt-based techniques for addressing the unique challenges of sarcasm detection in code-mixed languages and paves the way for future studies aimed at improving systems in multilingual social media contexts. We hope that this research will inspire continued exploration of advanced language models in tackling the diverse and evolving landscape of digital communication.

The evaluation of our sarcasm detection systems for Malayalam and Tamil languages provides valuable insights into the performance and challenges associated with handling these languages. For Malayalam, the system achieved a macro-F1 score of 0.50, which, while representing a moderate balance between precision and recall across classes, placed the system at the 13th position among competing models. This rank indicates that while our system performs adequately, there are several other models that have outperformed it in the specific context of Malayalam sarcasm detection.

In contrast, the Tamil sarcasm detection system achieved a higher macro-F1 score of 0.61, reflecting a better overall balance between precision and recall compared to the Malayalam system. Despite this improved score, the Tamil system was ranked 9th among competitors. This lower ranking, even with a higher F1 score, highlights the more competitive landscape for Tamil language tasks, where several systems have achieved similar or better performance levels.

These results underscore the varying degrees of difficulty in sarcasm detection across different languages and the importance of context in interpreting performance metrics like the macro-F1 score. The differences in ranking also suggest that the complexity of the language, the nature of the testing data, and the sophistication of competing models play significant roles in determining a system's success.

In future work, it will be crucial to explore ways to enhance the performance of both systems, particularly by focusing on the nuances of sarcasm in these languages, improving the quality and diversity of the training data, and leveraging more advanced modeling techniques. Additionally, a more in-depth analysis of the specific errors made by the systems could provide further insights into areas that require targeted improvements. Ultimately, the goal is to develop robust and competitive systems that can more effectively handle the challenges of sarcasm detection in both Malayalam and Tamil, contributing to the broader field of natural language processing for low-resource languages.
%\begin{acknowledgments}
%  Thanks to the developers of ACM consolidated LaTeX styles
%  \url{https://github.com/borisveytsman/acmart} and to the developers
%  of Elsevier updated \LaTeX{} templates
%  \url{https://www.ctan.org/tex-archive/macros/latex/contrib/els-cas-templates}.  
%\end{acknowledgments}

%%
%% Define the bibliography file to be used
\bibliography{sample-ceur}

\appendix


%The sources for the ceur-art style are available via
%\begin{itemize}
%\item \href{https://github.com/yamadharma/ceurart}{GitHub},
% \item \href{https://www.overleaf.com/project/5e76702c4acae70001d3bc87}{Overleaf},
%\item
%  \href{https://www.overleaf.com/latex/templates/template-for-submissions-to-ceur-workshop-proceedings-ceur-ws-dot-org/pkfscdkgkhcq}{Overleaf
%    template}.
%\end{itemize}

\end{document}